%% file: main.tex
\pdfoutput=1

\documentclass[11pt]{article}

\usepackage[final]{acl}

\usepackage{times}
\usepackage{latexsym}

\usepackage[T1]{fontenc}

\usepackage[utf8]{inputenc}

\usepackage{microtype}

\usepackage{inconsolata}

\usepackage{graphicx}
\usepackage{hyperref}       
\usepackage{url}            
\usepackage{booktabs}       
\usepackage{amsfonts}       
\usepackage{nicefrac}       
\usepackage{microtype}      
\usepackage{xcolor}         
\usepackage[inline]{enumitem}
\usepackage{dramatist}
\usepackage{dirtytalk}
\usepackage{multirow}
\usepackage{threeparttable}
\usepackage[linesnumbered,ruled,vlined]{algorithm2e}
\usepackage{graphicx}
\usepackage{subcaption}
\usepackage{caption}
\usepackage[linesnumbered,ruled,vlined]{algorithm2e}
\colorlet{verylightgray}{lightgray!99!white}
\usepackage{listings}
\title{MARCO: Multi-Agent Real-time Chat Orchestration}

\author{Anubhav Shrimal$^1$, Stanley Kanagaraj$^1$, Kriti Biswas$^1$, \\
        \textbf{Swarnalatha Raghuraman$^1$, Anish Nediyanchath$^1$,} \\
        \textbf{Yi Zhang$^2$ \and Promod Yenigalla$^1$} \\
        $^1$Retail Business Services, Amazon \\
        $^2$AWS Bedrock, Amazon \\
        \texttt{\{shrimaa, kstanly, kritibw, rswarnal, anishned,} \\
        \texttt{yizhngn, promy\}@amazon.com}}

\begin{document}
\maketitle
\input{content/01-abstract}

\input{content/02-introduction}
\input{content/03-approach/03-approach}
\input{content/04-experiments/04-experiments}
\input{content/05-results/05-results}

\input{content/06-conclusion}

\bibliography{references}

\input{content/09-appendix}

\end{document}

%% file: content/01-abstract.tex
\begin{abstract}
Large language model advancements have enabled the development of multi-agent frameworks to tackle complex, real-world problems such as to automate tasks that require interactions with diverse tools, reasoning, and human collaboration.
We present MARCO, a Multi-Agent Real-time Chat Orchestration framework for automating tasks using LLMs. MARCO addresses key challenges in utilizing LLMs for complex, multi-step task execution. It incorporates robust guardrails to steer LLM behavior, validate outputs, and recover from errors that stem from inconsistent output formatting, function and parameter hallucination, and lack of domain knowledge. Through extensive experiments we demonstrate MARCO's superior performance with 94.48\% and 92.74\% accuracy on task execution for Digital Restaurant Service Platform conversations and Retail conversations datasets respectively along with 44.91\% improved latency and 33.71\% cost reduction. We also report effects of guardrails in performance gain along with comparisons of various LLM models, both open-source and proprietary. The modular and generic design of MARCO allows it to be adapted for automating tasks across domains and to execute complex usecases through multi-turn interactions.
\end{abstract}

%% file: content/02-introduction.tex
\section{Introduction}

Advancements in LLM technology has led to a lot of interest in applying Agents framework to realise solutions which require complex interactions with the environment including planning, tools usage, reasoning, interaction with humans.
Recent works~\cite{Wang_2024-agents_survey1, huang2024understanding-agents_survey2} demonstrate potential of LLMs for creating autonomous Agents while there are numerous challenges to overcome and provide a seamless experience for end users who interact with the system at a daily basis.
LLMs are probabilistic next token prediction systems and by design, non-deterministic which can introduce inconsistencies in the output generation that can prove challenging for features like function calling, parameter value grounding, etc. There are also challenges of domain specific knowledge which can be an advantage and dis-advantage at the same time. 
LLMs have biases inherent in them which can lead to hallucinations, at the same time it may not have the right internal domain specific context which needs to be provided to get the expected results from an LLM.

We present our work on building a real time conversational task automation assistant framework with the following emphasis,
\begin{enumerate*}[label=(\arabic*)] 
    \item \textbf{Multi-turn Interface} for, 
    \begin{enumerate*}
        \item User conversation to execute tasks 
        \item Executing tools with deterministic graphs providing status updates, intermediate results and requests to fetch additional inputs or clarify from user.
    \end{enumerate*}
    \item \textbf{Controllable Agents} using a symbolic plan expressed in natural language task execution procedure (TEP) to guide the agents through the conversation and steps required to solve the task
    \item \textbf{Shared Hybrid Memory} structure, with Long term memory shared across agents which stores complete context information with Agent TEPs, tool updates, dynamic information and conversation turns.
    \item \textbf{Guardrails} for ensuring correctness of tool invocations, recover for common LLM error conditions using reflection and to ensure general safety of the system.
    \item \textbf{Evaluation} mechanism for different aspects and tasks of a multi-agent system.
\end{enumerate*}

This is demonstrated in the context of task automation assistant which supports adding usecase tasks to provide users a conversational interface where they can perform their intended actions, making it easier for them to refer to informational documents, interact with multiple tools, perform actions on them while unifying their interfaces.
We provide detailed comparison across multiple foundational LLMs as backbone for our assistant tasks like Claude Family models~\cite{claude3}, Mistral Family models~\cite{jiang2023mistral7b, jiang2024mixtral} and Llama-3-8B~\cite{llama3-8b} on Digital Restaurant Service Platform (DRSP) conversations and Retail conversations (Retail-Conv) datasets.
\input{content/02-1-relatedwork}

%% file: content/02-1-relatedwork.tex
\section{Related Work}~\label{appendix:related_work}
Improvements to LLM technology through the release of foundational LLMs like GPT-4~\cite{gpt4}, Claude~\cite{claude3} and Mixtral~\cite{jiang2024mixtral} has led to a flurry of research around autonomous agents and frameworks~\cite{Wang_2024-agents_survey1, huang2024understanding-agents_survey2}. 
Zero shot Chain-of-Thought (COT) reasoning \cite{kojima2023large} allows LLMs to perform task reasoning by making it think step by step. LLMs can invoke external tools based on natural language instructions. HuggingGPT \cite{shen2023hugginggpt} can coin series of model invocations to achieve complex tasks mentioned by the user. Toolformer \cite{schick2023toolformer} demonstrates how LLMs can be used as external tools through API invocations selecting the right arguments to be passed from few examples and textual instructions. Agents framework~\cite{zhou2023agents} discuss using natural language symbolic plans called (Standard Operating Procedures) SOPs which define transition rules between states as the agent encounters different situations to provide more control over agent behavior
along with memory to store relevant state information within the prompt~\cite{fischer2023reflective, rana2023sayplan} or long term context externally~\cite{zhu2023ghost, park2023generative}.
Amazon Bedrock Agents \footnote{\href{https://docs.aws.amazon.com/bedrock/latest/userguide/agents.html}{Amazon Bedrock Agents User Guide}} provide interface to quickly build, configure and deploy autonomous agents into business applications leveraging the strength of foundational models, while the framework abstracts the Agent prompt, memory, security and API invocations. 
LangGraph \footnote{\href{https://python.langchain.com/v0.2/docs/langgraph/}{LangGraph library}} is an extension of LangChain which facilitates the creation of stateful, multi-actor applications using large language models (LLMs) by adding cycles and persistence to LLM applications thus enhancing their Agentic behavior. It coordinates and checkpoints multiple chains (or actors) across cyclic computational steps.  
While these frameworks present novel ways for LLMs to act in a desired behaviour, they often have accuracy-latency trade-off where to improve on the accuracy the system latency increases due to multi-step planning and thinking~\cite{react,cot}. Our proposed solution, MARCO, not only interacts with user in a multi-turn fashion but also has multi-turn conversation with deterministic multi-step functions which comprises of pre-determined business logic or task execution procedure (TEP) requiring agents only at intelligent intervention related steps. Along with the usecase TEPs, multi-step functions and robust guardrails to steer LLM behaviour, MARCO is able to perform complex tasks with high accuracy in less time as detailed in subsequent sections.

%% file: content/03-approach/03-approach.tex
\section{MARCO: Multi-Agent Real-time Chat Orchestration} \label{sec:approach}
In this section, we discuss our approach for MARCO. Section~\ref{sub_sec:approach-problem_statement} formulates the problem statement in terms of Task Automation via real-time chat, followed by components of MARCO in section~\ref{sub_sec:approach-marco_components}
and the evaluation methods on performance and latency for MARCO in section~\ref{sub_sec:approach-evaluation_methods}.

\input{content/figures/marco_usecase_flow}
\input{content/03-approach/03-1-approach-problem_statement}

\input{content/03-approach/03-2-approach-marco_components/03-2-approach-marco_components}
\input{content/03-approach/03-3-approach-dataset_preparation}
\input{content/03-approach/03-4-approach-evaluation_methods}


%% file: content/figures/marco_usecase_flow.tex
\begin{figure*}[htbp]
  \centering
  \includegraphics[width=\textwidth]{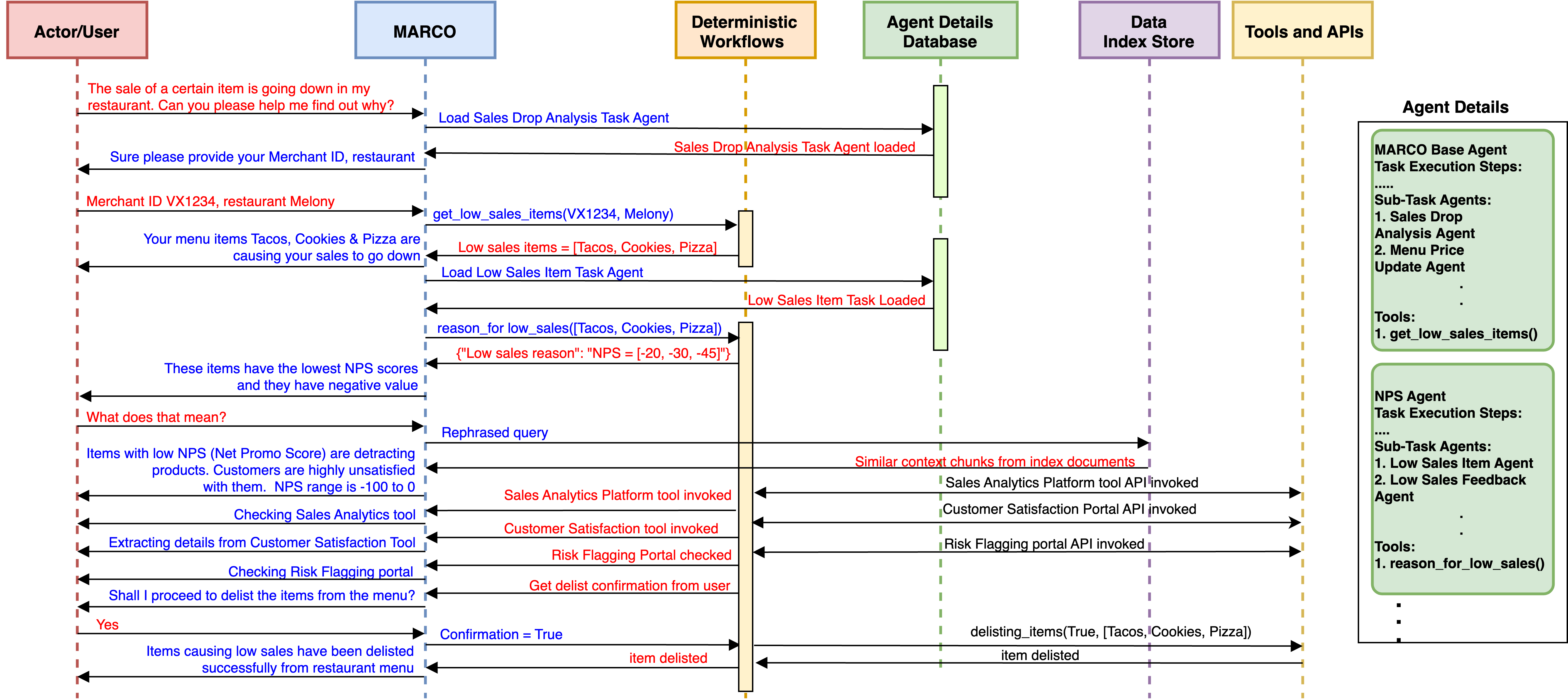}
  \caption{Multi-Agent Conversation Flow in MARCO Framework. This diagram illustrates the complex interactions within the \emph{MARCO} system as it addresses a user's query about declining sales. It showcases MARCO's orchestration of multiple components including the MARCO Base agent, specialized task agents, deterministic multi-step workflows, data stores, and external tools/APIs. The figure demonstrates MARCO's capability to manage multi-turn communications with both the user and various system components, highlighting its process of task decomposition, information gathering, analysis, and action execution in response to real-world business scenarios.}
  \label{fig:marco_usecase_flow}
\end{figure*}

%% file: content/03-approach/03-1-approach-problem_statement.tex
\subsection{Problem Statement} \label{sub_sec:approach-problem_statement}
Given an  user (\emph{Actor}), 
who wishes to perform a task with intent \emph{I}~$\in$~\{\emph{OOD, Info, Action}\}; where \emph{Out-Of-Domain (OOD)} intent is defined as any user query which is not in scope of the system such as malicious query to jailbreak~\cite{shen2023do-jailbreak1, rao2024tricking-jailbreak2} the system, foul language or unsupported requests, \emph{``Info''} intent is defined as getting information from predefined data-sources and indexed documents ($D_{index}$), and \emph{``Action''} intent is defined as a performing a usecase related task ($u_x$) which involves following a series of instructions/steps (Task Execution Procedure, $TEP_x$) defined for the usecase and accordingly invoking the right set of tools/functions ($F_*^x = \{F_1^x, F_2^x, ..., F_n^x\}$) with the identified required parameters ($P_*^x = \{P_{F_1^x}, P_{F_2^x}, ..., P_{F_n^x}\}$) for each function respectively. The objective for a
task automation system is to, 
\begin{enumerate*}[label=(\arabic*)] 
    \item interpret the user intent \emph{I} for each query,
    \item identify the relevant usecase $u_x$,
    \item understand the steps mentioned in its $TEP_x$, 
    \item accordingly call the right sequence of tools $F_*^x$ with required parameters $P_*^x$,
    \item correlate $TEP_x$, tool responses and requirements and conversation context to communicate back with the user, and
    \item be fast and responsive for a real-time chat.
\end{enumerate*}

An example scenario is shown in figure~\ref{fig:marco_usecase_flow} where User first asks ``\emph{The sale of certain item is going down in my restaurant. Can you please help me find out why?}'', i.e. \emph{I}~$=$~\emph{Action} for which MARCO then loads the agent with TEP for Sales Drop Analysis usecase ($TEP_{sd}$) and then goes on to call relevant function \emph{F}$=$[\emph{get\_low\_sales\_item}, \emph{reason\_for\_low\_sales}] with respective required parameter values \emph{merchant\_id} and \emph{restaurant\_name}. It is worth noting that the interaction with MARCO is multi-turn, both with the user as well as the functions being called where the functions may provide intermediate communications or ask for information to proceed further (for example, \emph{confirmation=True}).

%% file: content/03-approach/03-2-approach-marco_components/03-2-approach-marco_components.tex
\subsection{MARCO -- Components} \label{sub_sec:approach-marco_components}
MARCO built for task automation has 4 primary LLM components, (i) Intent Classifier, (ii) Retrieval Augmented Generation (RAG) to answer domain related informational queries, (iii) MARS for tasks orchestration and execution, and (iv) Guardrails. The sections below cover each of the component, except for RAG where the implementation details are out of scope for this paper.

\input{content/03-approach/03-2-approach-marco_components/03-2-1-marco_components-intent_classifier}
\input{content/03-approach/03-2-approach-marco_components/03-2-2-marco_components-mars}
\input{content/03-approach/03-2-approach-marco_components/03-2-4-marco_components-guardrails}
\input{content/03-approach/03-2-approach-marco_components/03-2-5-marco_components-context_sharing}

%% file: content/03-approach/03-2-approach-marco_components/03-2-1-marco_components-intent_classifier.tex
\subsubsection{Intent Classifier} \label{subsub_sec:approach-intent_classifier}
Intent Classifier's (IC) primary role is to understand the intent behind an incoming user message considering the conversation context, and to seamlessly orchestrate between RAG for answering informational queries and Multi-Agents system (MARS) to execute supported tasks. 
IC also takes the role of first level guardrails to identify and gracefully reject queries to protect the underlying modules from harmful jailbreak instructions and \emph{Out-Of-Domain} (OOD) queries. At a high level IC performs intent classification into one of the three supported classes \emph{\{Info, Action, OOD\}}, leveraging language understanding capability of LLMs. Major challenges faced by intent classifier can be found in Appendix~\ref{appendix:intent_classifier}.

%% file: content/03-approach/03-2-approach-marco_components/03-2-2-marco_components-mars.tex
\subsubsection{Multi-Agent Reasoner and Orchestrator (MARS)} \label{subsub_sec:approach-mars} 
When a user query is classified as an \emph{I = Action} intent, the chat conversation history is redirected to \emph{MARS} (Multi-Agent Reasoner and Orchestrator) module which is a Multi-Agent system responsible for
\begin{enumerate*}[label=(\arabic*)]
    \item understanding the user's request and tool responses in the chat context
    \item planning and reasoning for the next action according to the Task Execution Procedure (TEP) steps,
    \item selecting relevant LLM Agent for the task, and
    \item invoking the relevant tools/tasks with their required parameters.
\end{enumerate*}
The key component of MARS are the LLM Agents, which we call \emph{Task Agents}. These Task Agents comprise of their own TEP steps, tools/functions also known as \emph{Deterministic Tasks}, \emph{Sub-Task-Agents} (dependent Task Agents) and common instructions for reasoning and output formatting. We will explain each of these in detail:

\input{content/figures/sop_agents_dag}

\textbf{Deterministic Tasks:} Task Execution Procedure (TEP) steps can be very complex with multiple instructions and steps to follow based on a given usecase scenario.
While some of these steps require high judgement and reasoning (understanding natural language to parse required arguments, intents, performing checks defined in plain text without writing explicit code), most of the steps in the TEP are deterministic sequence of API calls, processing and propagating the output gathered from API$_1$ to API$_2$ and so on. Such sequence of deterministic steps can be encapsulated as a single tool to the LLM Agent, which when called performs the sequence of these deterministic steps and communicates with the agent intermittently with updates or any high judgement reasoning or inputs required by the underlying APIs (for example refer to Appendix~\ref{appendix:retail_conv_dataset}). 

\textbf{Task Agents:} A usecase TEP can be divided into multiple Sub-Tasks which are logical abstractions of complex steps inside the TEP. For example, if a usecase $u_x$ has sub-branches \{a, b, c\}, each with their own set of steps to follow, then each can be created as a Task Agent ($A$) where Agent $A_x$ has agents $\{A_a, A_b, A_c\}$ as it's child Task-Agents. Each of these child Agents may further have their own children Agents based on their TEP complexity. A Task Agent has the steps comprised in its TEP along with the list of available determinisitic tasks/functions that the particular Agent can utilize, for e.g. The ``Sales Drop Analysis'' usecase Agent ($A_{sd}$) may have a function named \emph{get\_low\_sales\_items(merchant\_id, restaurant\_name)} function but will not have \emph{update\_menu\_item\_price(menu\_item, price)} function as it is not a valid dependency. An Agent also has the information of the immediate child Sub-Task-Agents in its hierarchy so that it can invoke another child agent if required during its planning. Figure~\ref{fig:sop_agents_dag} shows an example multi-agent hierarchy for DRSP dataset where \emph{MARCO Base Agent}, using which the system is first initialized, is the main agent with its own TEP steps, tools and Sub-Agents i.e. the usecases which are added onto the platform. 

Agent's LLM input prompt has sub-agents, tools, reasoning and formatting instructions and chat history embedded using which it has to auto-regressively generate the output. We prompt the underlying LLM to generate the \say{\emph{message}} which is to be conveyed to the \emph{Actor} and the corresponding \say{\emph{action}} which could be to invoke a deterministic task with the arguments the Agent provides or switching to a child Task-Agent. 
Appendix~\ref{appendix:llm_input_output_formatting} provides more details on input and output formatting.

%% file: content/figures/sop_agents_dag.tex
\begin{figure}[htbp]
  \centering
  \includegraphics[width=0.98\columnwidth]{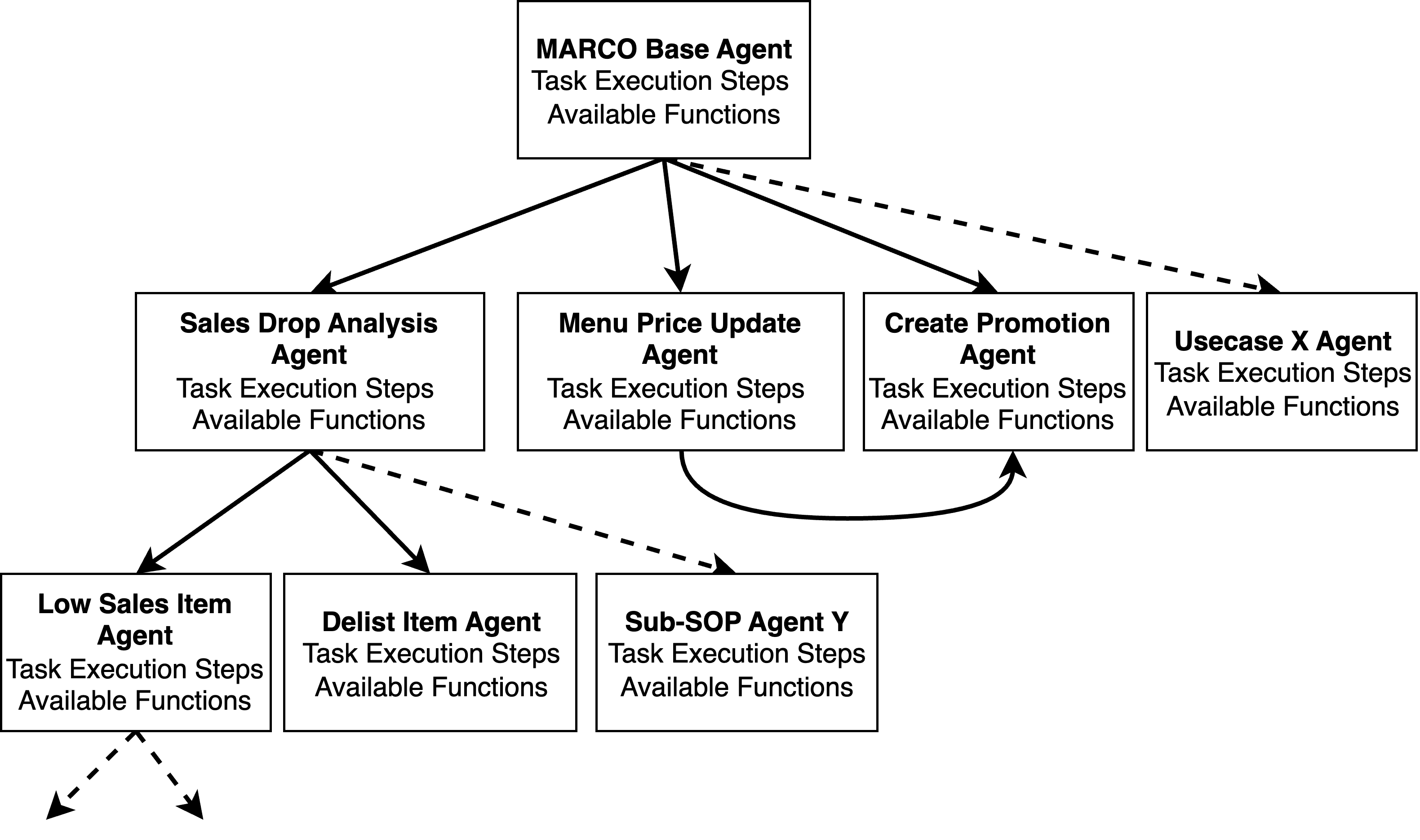}
  \caption{Multi-Agents Hierarchy example for Digital Restaurant Service Platform dataset. A directed acyclic graph in which each agent has it's own Task Execution Procedure (TEP) steps, functions and dependent Sub-Task Agents.}
  \label{fig:sop_agents_dag}
\end{figure}

%% file: content/03-approach/03-2-approach-marco_components/03-2-4-marco_components-guardrails.tex
\subsubsection{Guardrails} \label{subsub_sec:approach-guardrails}
LLMs exhibit stochastic behavior, generating varying outputs for the same input. They are susceptible to hallucination~\cite{bang2023multitask-hallucination1, guerreiro2023hallucinations-2}, producing responses with fabricated or inaccurate information. It is crucial to establish mechanisms to steer LLMs in the desired direction for reliable systems. We introduce guardrails to identify issues and prompt the LLM-Agents to reflect on their mistakes, correcting their responses. Common issues and proposed guardrail solutions are:
\begin{enumerate*}[label=(\arabic*)] 
    \item \textbf{Incorrect Output Formatting:} Generating incorrect formats despite detailed instructions, causing parsing issues. If parsing fails, a reflection prompt is added to the Agent's chat history, and sent for a retry. 
    \item \textbf{Function Hallucination:} Hallucinating non-existent function names, even when prompted to use only existing tools. Our guardrails checks if the generated function name exists in the available tools and Sub-Agents. If not, reflection prompt is added.
    \item \textbf{Function Parameter Value Hallucination:} When making function calls with required parameters, LLMs sometimes hallucinate parameter values instead of asking relevant questions to the user. This often occurs due to pre-trained dataset biases because they have seen this pattern frequently during pre-training, making it challenging to unlearn using prompting techniques. For each function parameter $p$, the module checks if $p$ is part of the function schema; if not, $p$ is removed (e.g., for \emph{get\_low\_sales\_items(merchant\_id, restaurant\_name)}, the Agent also generated \emph{menu\_item} as a parameter). The parameter value for non-boolean parameters is grounded to be present in the \emph{Actor} message history; if not, it is classified as hallucination (e.g., \emph{Actor} said, \say{update menu price of item X to \$50} and Agent generated \emph{marketplace=\say{US}} which was not mentioned by the Actor).
    \item \textbf{Lack of Domain Knowledge:} Although pre-trained LLMs possess good general world knowledge, they may lack certain domain-specific knowledge, especially in lesser-known domains.
    We define a list of static rules for each parameter based on the type, constituent values, length and more (e.g., \say{merchant\_id value has a minimum\_length=6 and maximum\_length=8, is an alphanumeric string}). The guardrails module checks if the generated value satisfies these rules; if not, a reflection prompt with rule failures is added. Parameter properties and definitions are also introduced in the Agent prompt as \emph{<helpful\_definitions>...</helpful\_definitions>} to provide explicit in-domain knowledge for e.g., \say{<helpful\_definitions>merchant\_id is 6-8 character alphanumeric string, restaurant\_code is 4-5 character alphanumeric string</helpful\_definitions>}, which helps the Agent to then disambiguate these values when provided without names by the \emph{Actor} (e.g. \say{VX1234, BL123}).
\end{enumerate*}
The number of retries with reflection is limited to 2 (\emph{NUM\_RETRIES=2}) for real-time chat system latency. Appendix Algorithm~\ref{algo:marco_reflection_guardrails} provides detailed flow of guardrails.

%% file: content/03-approach/03-2-approach-marco_components/03-2-5-marco_components-context_sharing.tex
\subsubsection{Context Sharing} \label{subsub_sec:approach-context_sharing}
As MARCO has multiple components (IC, MARS, RAG) and is a multi-turn multi-agent conversation system, it needs a mechanism to share the context amongst each other. Along with the usual roles of \emph{[[SYSTEM], [USER], [AGENT]]} similar to Bedrock's Claude messages API format\footnote{\href{https://docs.aws.amazon.com/bedrock/latest/userguide/model-parameters-anthropic-claude-messages.html}{Bedrock Claude messages API documentation}}, we introduce separate roles for function responses and guardrails, \emph{[FUNCTION\_RESPONSE], [GUARDRAILS]}, respectively. This allows LLM-Agents to better differentiate each message in the chat history as the conversation is multi-turn from both \emph{Actor} and \emph{Deterministic tasks} (for example Figure~\ref{fig:marco_usecase_flow} \emph{reason\_for\_low\_sales()} task communicates multiple times to MARCO), and it prevents jailbreaking by malicious \emph{Actors}. When a Parent Agent (\emph{Agent}$_p$) loads its Child Agent (\emph{Agent}$_c$), the \emph{[SYSTEM]} prompt is updated with \emph{Agent}$_c$ details and a message is added to the chat history to capture that an agent switch has occurred. The common chat history thread is shared among all Task-Agents for a chat session, as any information provided to \emph{Agent}$_p$ by the \emph{Actor} or a \emph{function response} might be useful for the executing \emph{Agent}$_c$'s task execution procedure (TEP) steps. 

%% file: content/03-approach/03-3-approach-dataset_preparation.tex

%% file: content/03-approach/03-4-approach-evaluation_methods.tex
\subsection{Evaluation Methods} \label{sub_sec:approach-evaluation_methods}
A real-time
task automation system should have highly accurate execution as well as fast turn-around time. Keeping these tenets in mind, we evaluate MARCO components on quality and accuracy of generated responses along with time taken to produce such outputs. 
For evaluating \emph{MARS} we compare the expected function call and parameter ($F_i^x, P_{F_i^x}$) with the generated function call and parameter ($\hat{F}_i^x, \hat{P}_{F_i^x}$) whenever an action is expected in test data. We also implemented an LLM response evaluation prompt which takes in two response messages ($m_1$, $m_2$) and returns \emph{True} if semantics of $m_1$ and $m_2$ are the same else \emph{False}. An manual audit based evaluation is also performed to validate the efficacy of our LLM response evaluation prompt
(LLM evaluation prompt detailed in Appendix~\ref{appendix:mars_llm_evaluation_prompt}). 
Both, the generated function call and response message semantics, should be evaluated as correct with the ground truth to mark the complete generated output as valid. We calculate the accuracy as the number of test cases where MARS's complete generated output is valid.
For each component we also calculate and compare the latency and cost of response generation as it is a real-time chat system.

%% file: content/04-experiments/04-experiments.tex
\section{Experimental Setup}
\textbf{Dataset:} For our experiments, we curated two conversational orchestration test datasets, Digital Restaurant Service Platform (DRSP-Conv) and Retail-Conv, each with 221 and 350 multi-turn conversations in the restaurant services and retail services domain respectively. These conversations are a mix of \emph{Out-Of-Domain (OOD), Action} and \emph{Info} queries with multi-turn interactions with both, User and Deterministic Tasks (an example conversation flow in the dataset is shown in figure~\ref{fig:marco_usecase_flow} for DRSP-Conv). The dataset covers usecases along with their natural language Task Execution Procedure (TEP) steps, supported functions (deterministic tasks and utility tools~\footnote{utility tools are simple functions to get specific data, e.g., get\_menu\_item\_name(), get\_menu\_item\_price(), etc.}) and sub-task agents.
Each test conversation has multiple \emph{Assistant} (Agent) messages (replying to the user, loading an agent, calling a deterministic task, or answering an informational query). 
We use these datasets for evaluating MARCO on the defined performance metrics. Hyper-parameter details mentioned in Appendix~\ref{appendix:hyperparams}.

\textbf{Baseline:} We implement MARCO with a single agent-based prompt as a baseline to compare with our multi-agent proposed solution, on performance, latency and cost. To achieve this, the usecase sub-task TEP steps in the datasets were combined into the parent agent TEP steps to create a single agent TEP and the datasets were modified accordingly to support the single agent baseline. 


%% file: content/05-results/05-results.tex
\input{content/tables/marco_llms_compare}

\section{Experiments \& Results}
In this section we detail the various experiments to evaluate our proposed solution, MARCO, on task specific performance, operational performance (latency, run-time cost), scalability and ablations.

\input{content/figures/cost_analysis}
\textbf{MARS Operational Performance:} 
We compare the accuracy, latency of \emph{MARS} (Multi-Agent Reasoner and Orchestrator) using various open-source (llama-3-8B, mistral-7B, mixtral-8x7B) and proprietary instruction-tuned LLMs (claude-instant-v1, claude-v2.1, claude-v3-haiku, and claude-v3-sonnet) in Table~\ref{tab:marco_llms_compare}. We observe that \emph{claude-3-sonnet} performs best with 94.48\% and 92.74\% accuracy and 5.61 and 5.85 seconds latency including all reflection guardrails for DRSP-Conv and Retail-Conv datasets respectively. Sonnet is also ~30\% faster and 60\% cheaper than \emph{claude-v2.1}, making it cost-effective as shown in figure~\ref{fig:mars_cost} for MARCO implementation costs using various LLMs assuming 5000 requests with average input and output tokens calculated empirically (refer Appendix~\ref{appendix:cost_analysis}). Open-source LLMs underperform even with reflection guardrails, suggesting the need for fine-tuning as future work. We found a Cohen's kappa~\cite{cohens_kappa} of 0.65 (96.66\% agreement) between human auditors and our LLM semantics similarity matching prompt for evaluation, indicating a high level of agreement. Intent Classifier 
has 94.53\% using \emph{claude-v3-sonnet} 3-class classification accuracy with a latency of 1.98 seconds (refer Appendix Table~\ref{tab:intent_classifier_llm_compare}). 

\input{content/tables/baseline_marco_compare}

\textbf{Single-Agent Baseline vs Multi-Agent (MARS) performance:}
Through this experiment, we aim to demonstrate the effectiveness of MARS against a Single-Agent baseline covering all usecases. Table~\ref{tab:baseline_marco_compare} shows that our proposed multi-agent system, MARCO, outperform single-agent baseline by +11.77\% and +4.36\% with all guardrails included on respective datasets. 
Also, the latency of Single-Agent baseline is on average 44.91\% higher and increases the cost by 33.71\% (\$70.29 per 5k requests) compared to MARS (\$52.57 per 5k requests) due to longer prompt length for the Agent.

\input{content/figures/mars_reflection_vs_only_retry}
\textbf{Effects of Reflection Guardrails:}
Through this experiment, we compare MARS's performance when all reflection guardrails, as discussed in section~\ref{subsub_sec:approach-guardrails}, are added vs without adding any guardrails. As shown in table~\ref{algo:marco_reflection_guardrails} adding reflection guardrails provides a +28.14\% and +31.85\% boost in accuracy while increasing the latency only by 1.54 and 1.24 seconds on average for DRSP-Conv and Retail-Conv respectively. Figure~\ref{fig:mars_reflection_vs_only_retry} illustrates the impact of our proposed reflection guardrails, where the first retry with reflection resolves all but two errors, whereas without any reflection prompt (using the original prompt on retires), error rates remain high even after four retries. 
Appendix~\ref{appendix:reflection_guardrail_ablation} shows the effects of removing each reflection type. On further deep dive we observe that claude-3-haiku has better performance than larger counterparts (claude-3-sonnet and claude-v2.1) when no guardrails are applied primarily due to its effectiveness in following output formatting instructions and generating correct outputs more often. Hence Haiku could be a viable option when cost of retries and latency have to be reduced further.

\textbf{Effects of Temperature, Input \& Output Token Lengths:}
Increasing temperature hyper-parameter allows an LLM to be more creative while generating a response. We observed that setting the value \emph{temparature=0} gives the best accuracy for MARS (Appendix Table~\ref{fig:mars_temperature_effect}), which is understandable as Task Execution Procedure (TEP) instruction following and function calling should be reliable and should not vary. Also, with increasing number of input and output tokens, the latency of MARCO increases (Appendix~\ref{appendix:temperature_and_latency}).



%% file: content/tables/marco_llms_compare.tex
\begin{table}[htbp]
\centering
\resizebox{\columnwidth}{!}{%
\begin{tabular}{@{}lcccc@{}}
\toprule
\multicolumn{5}{c}{DRSP-Conv dataset}                                                                                                  \\ \midrule
\textbf{}             & \multicolumn{2}{c}{With All Reflection Guardrails (retries=2)} & \multicolumn{2}{c}{No Guardrails (retries=0)} \\ \cmidrule(lr){2-2} \cmidrule(lr){4-4}
Model Name            & Accuracy (\%) ± Std dev            & Latency (secs)            & Accuracy (\%) ± Std dev    & Latency (secs)   \\ \midrule
llama-3-8b-instruct   & 42.44 ± 2.01                       & 3.75                      & 15.93 ± 0.98               & \textbf{1.9}     \\
mistral-7b-instruct   & 66.33 ± 1.04                       & 4.92                      & 59.28 ± 1.06               & 2.9              \\
mixtral-8x7b-instruct & 40.64 ± 1.51                       & 17.77                     & 32.67 ± 0.38               & 15.55            \\
claude-instant-v1     & 74.38 ± 1.4                        & 3.25                      & 53.12 ± 3.83               & 2.53             \\
claude-3-haiku        & 84.8 ± 0.88                        & \textbf{2.14}             & \textbf{75.2 ± 0.87}       & 2.24             \\
claude-v2.1           & 88.51 ± 0.76                       & 8.44                      & 64.52 ± 1.04               & 6.61             \\
claude-3-sonnet       & \textbf{94.48 ± 0.59}              & 5.61                      & 66.34 ± 0.82               & 4.07             \\ \midrule
\multicolumn{5}{c}{Retail-Conv dataset}                                                                                                \\
\textbf{}             & \multicolumn{2}{c}{With All Reflection Guardrails (retries=2)} & \multicolumn{2}{c}{No Guardrails (retries=0)} \\ \cmidrule(lr){2-2} \cmidrule(lr){4-4}
Model Name            & Accuracy (\%) ± Std dev            & Latency (secs)            & Accuracy (\%) ± Std dev    & Latency (secs)   \\ \midrule
llama-3-8b-instruct   & 49.68 ± 1.55                       & 3.44                      & 17.82 ± 1.12               & \textbf{1.64}    \\
mistral-7b-instruct   & 55.32 ± 0.77                       & 4.89                      & 50.72 ± 0.66               & 3.06             \\
mixtral-8x7b-instruct & 48.31 ± 0.60                       & 12.94                     & 40.49 ± 0.93               & 5.96             \\
claude-instant-v1     & 76.61 ± 0.81                       & 4.14                      & 60.56 ± 0.24               & 2.94             \\
claude-3-haiku        & 87.82 ± 0.44                       & \textbf{2.45}             & 77.66 ± 1.01               & 2.43             \\
claude-v2.1           & 92.34 ± 0.49                       & 8.2                       & \textbf{78.87 ± 0.61}      & 6.95             \\
claude-3-sonnet       & \textbf{92.74 ± 0.49}              & 5.85                      & 60.89 ± 0.81               & 4.61             \\ \bottomrule
\end{tabular}%
}
\caption{LLMs performance comparison for MARCO with and without guardrails on DRSP-Conv and Retail-Conv datasets averaged across 5 runs.}
\label{tab:marco_llms_compare}
\end{table}

%% file: content/figures/cost_analysis.tex
\begin{figure}[ht]
    \centering
    \begin{subfigure}[b]{0.48\columnwidth}
        \centering
        \includegraphics[width=\textwidth]{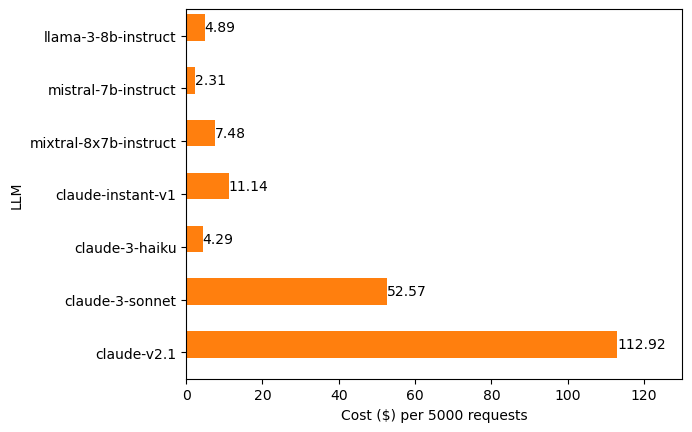}
        \caption{Cost of MARS.}
        \label{fig:mars_cost}
    \end{subfigure}
    \hfill
    \begin{subfigure}[b]{0.48\columnwidth}
        \centering
        \includegraphics[width=\textwidth]{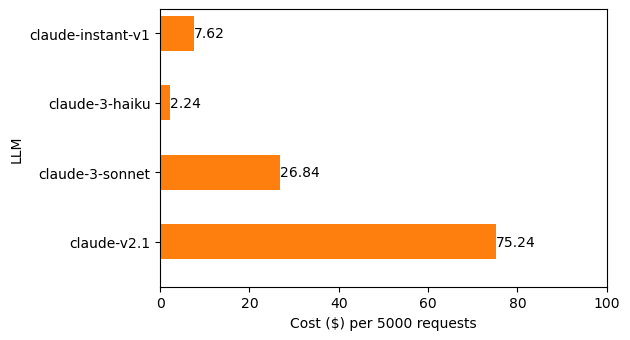}
        \caption{Cost of Intent Classifier.}
        \label{fig:ic_cost}
    \end{subfigure}
    \caption{Cost (\$) of MARCO components for every 5000 requests using various LLMs.}
    \label{fig:marco_cost_analysis}
\end{figure}

%% file: content/tables/baseline_marco_compare.tex
\begin{table}[htbp]
\centering
\resizebox{\columnwidth}{!}{%
\begin{tabular}{@{}lcccc@{}}
\toprule
\multicolumn{5}{c}{DRSP-Conv dataset}                                                                                                                                                            \\
                                                                                & \multicolumn{2}{c}{With All Reflection Guardrails (retries=2)} & \multicolumn{2}{c}{No Guardrails (retries=0)} \\ \cmidrule(lr){2-2} \cmidrule(lr){4-4}
Model Name                                                                      & Accuracy (\%) ± Std dev            & Latency (secs)            & Accuracy (\%) ± Std dev    & Latency (secs)   \\ \midrule
\begin{tabular}[c]{@{}l@{}}MARCO Single-Agent \\ (claude-3-sonnet)\end{tabular} & 82.71 ± 0.68                       & 6.89                      & \textbf{70.63 ± 2.77}      & 5.72             \\
\begin{tabular}[c]{@{}l@{}}MARCO Multi-Agent\\ (claude-3-sonnet)\end{tabular}   & \textbf{94.48 ± 0.59}              & \textbf{5.61}             & 66.34 ± 0.82      & \textbf{4.07}    \\ \midrule
\multicolumn{5}{c}{Retail-Conv dataset}                                                                                                                                                          \\
                                                                                & \multicolumn{2}{c}{With All Reflection Guardrails (retries=2)} & \multicolumn{2}{c}{No Guardrails (retries=0)} \\ \cmidrule(lr){2-2} \cmidrule(lr){4-4}
Model Name                                                                      & Accuracy (\%) ± Std dev            & Latency (secs)            & Accuracy (\%) ± Std dev    & Latency (secs)   \\ \midrule
\begin{tabular}[c]{@{}l@{}}MARCO Single-Agent \\ (claude-3-sonnet)\end{tabular} & 88.38 ± 0.90                       & 9.77                      & \textbf{80.07 ± 0.77}      & 8.81             \\
\begin{tabular}[c]{@{}l@{}}MARCO Multi-Agent\\ (claude-3-sonnet)\end{tabular}   & \textbf{92.74 ± 0.49}              & \textbf{5.85}             & 60.89 ± 0.81               & \textbf{4.61}    \\ \bottomrule
\end{tabular}%
}
\caption{Comparing MARCO single-agent and multi-agent with and without guardrails on DRSP-Conv and Retail-Conv datasets averaged across 5 runs.}
\label{tab:baseline_marco_compare}
\end{table}

%% file: content/figures/mars_reflection_vs_only_retry.tex
\begin{figure}[ht]
\centering
\includegraphics[width=0.98\columnwidth]{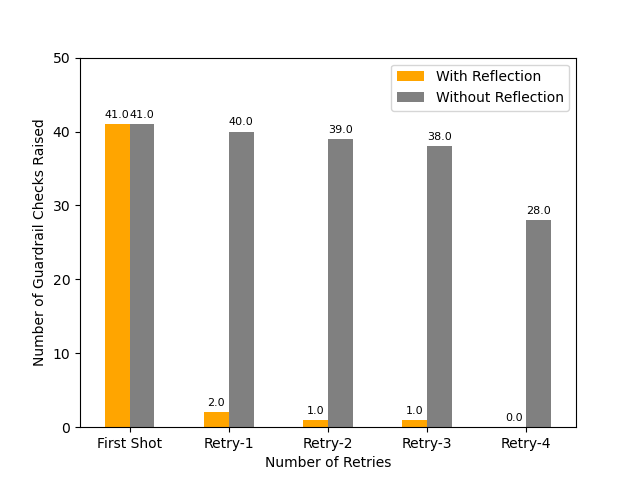}
\captionof{figure}{Impact of Reflection Prompts on Guardrail Error Recurrence During Retries. This graph compares the number of guardrail errors persisting across multiple retry attempts, with and without the use of reflection prompts. It demonstrates that incorporating reflection prompts significantly reduces error recurrence, typically resolving issues within the first retry. In contrast, retrying without reflection shows a gradual decrease in errors but fails to eliminate them entirely even after four attempts.}
\label{fig:mars_reflection_vs_only_retry}
\end{figure}

%% file: content/06-conclusion.tex
\section*{Conclusion}
We presented MARCO, a multi-agent real-time chat orchestration framework for automating
tasks using large language models (LLMs) addressing key challenges in utilizing LLMs for complex, multi-step task execution with high accuracy and low latency including reflection guardrail prompts for steering LLM behaviour and recover from errors leading to +30\% accuracy improvement. We demonstrated MARCO's superior performance with up to +11.77\% and +4.36\% improved accuracy against single agent baseline for two datasets, DRSP-Conv and Retail-Conv, and improved latency by 44.91\% and 33.71\% cost reduction. The modular and generic design of MARCO allows it to be adapted for automating tasks across various domains
wherever complex tasks need to be executed through multi-turn interactions using LLM-powered agents. 

%% file: content/09-appendix.tex
\appendix
\section{Appendix}

\subsection{Discussion}
\label{sub_sec:approach-discussion}
While as part of this work our experiments are focused towards Digital Restaurant Service and Retail task automations, the design for MARCO is generic LLM Agents based framework and can be adapted to any domain where the system is required to follow standard task execution steps to solve for a usecase while using set of available tools and interacting with an end user. Also, the guardrails and evaluation methods are generic for such a framework. As Intent Classifier, RAG and MARS are independent modules, we execute them in parallel to reduce the latency of our real-time chat system. The output from MARS or RAG is picked according to IC's classification. 

\input{content/algorithms/guardrails_algorithm}
\input{content/tables/marco_reflection_ablation}

\subsection{Hyper-parameters:}~\label{appendix:hyperparams}
For all experiments, unless specified otherwise, we used the underlying LLM as \emph{claude-3-sonnet} with \emph{temperature=0}, \emph{max\_output\_tokens=1000} and \emph{Top-P}, \emph{Top-K} values as defaults. We use LLM APIs provided by Amazon Bedrock dated July 1, 2024  for output generation. The maximum number of retires on any guardrail failure was set to 2, and if the issue still persisted, a constant \say{Facing Technical Issue} response was sent back. We ran each experiment five times and published the average and standard deviation for the results. We publish the cost calculation numbers with AWS Bedrock pricing~\footnote{\href{https://aws.amazon.com/bedrock/pricing/}{Bedrock API Pricing Documentation}} in this work.

\subsection{Reflection Guardrails Ablation}~\label{appendix:reflection_guardrail_ablation}
Table~\ref{tab:marco_reflection_ablation} performs an ablation of each of the reflection prompts discussed in section~\ref{subsub_sec:approach-guardrails}. The results show that each reflection prompt contributes to the performance enhancement of MARCO without which the performance drops significantly on DRSP-Conv and Retail-Conv datasets. The latency also does not increase much due to re-trying with reflection with an average increase of only 1.54 and 1.24 seconds respectively when adding all guardrails to the system in \emph{claude-3-sonnet}.

\input{content/figures/mars_temperature_effect}
\subsection{Effects of Temperature, Input \& Output Token Lengths:}~\label{appendix:temperature_and_latency}
\textbf{Effects of Temperature:} We vary the temperature hyper-parameter at an increment of +0.2 from 0 to 1 and compare the performance accuracy of MARS using \emph{claude-3-sonnet} and \emph{claude-v2.1}. The results suggest that \emph{temperature=0} performs the best for MARCO.

\input{content/figures/mars_input_output_token_latency_relation}
\textbf{Effects of Input and Output Tokens on Latency:} In figure~\ref{fig:mars_input_output_token_latency_relation} we plot the latency of MARS using \emph{claude-3-sonnet} with respect to input tokens (x-axis). We further color code each instance on the plot based on the number of output tokens generated within a given range. The results show a correlation between the growing number of input tokens leading to an increase in the latency while also having large number of output tokens for similar input token length leading to further increase in the latency.

\subsection{Cost Analysis} \label{appendix:cost_analysis}
To calculate the cost of various LLM version we assume that the task automation system has on average:
\begin{enumerate}
    \item $100$ active users per day,
    \item $50$ messages per chat,
    \item $X$ input tokens per LLM request (calculated empirically from our experiments in table~\ref{tab:marco_llms_compare}),
    \item $Y$ output tokens per LLM request (calculated empirically from our experiments in table~\ref{tab:marco_llms_compare}).
    \item $\$Z_i / 1000$ input tokens and $\$Z_o / 1000$ output tokens cost of LLM API invocation.
\end{enumerate}
Then the cost of the system ($C$) in product to serve 5k requests ($100 * 50 = 5000$) is calculated as follows:
\begin{equation}
    C = (5000 * X * Z_i / 1000) + (5000 * Y * Z_o / 1000)
\end{equation}
Single-Agent baseline has on an average 3946 input and 148 output tokens which leads to a total of \$70.29 per 5k requests cost using \emph{claude-3-sonnet}.\footnote{\href{https://aws.amazon.com/bedrock/pricing/}{Pricing of Bedrock API Documentation}} Pricing for MARCO components (IC and MARS) for various LLMs is shown in figure~\ref{fig:marco_cost_analysis}. The results state that using \emph{claude-v2.1} is 2.14 times costly compared to \emph{claude-3-sonnet}. Similarly, for Intent Classifier using \emph{claude-3-sonnet} followed by \emph{claude-instant-v1} is an ideal choice to keep latency and cost in mind while also comparing the performance (refer table~\ref{tab:intent_classifier_llm_compare}).

\input{content/tables/intent_classifier_prompting_techniques}
\input{content/tables/intent_classifier_llm_compare}
\subsection{Intent Classifier prompting techniques}~\label{appendix:intent_classifier}
In this section we explain the various prompting techniques that we employed to improve the performance of Intent Classifier. The primary objective of the Intent Classifier is to classify between \emph{I=Info, I=Action} intents, while also adeptly managing casual conversational contexts such as greetings, out-of-domain inquiries, and potential jailbreak attempts. Major challenges that we have addressed for IC are:
\begin{enumerate}
    \item Disambiguate closely related queries that can have different meaning and should be classified accordingly. For e.g., \say{What is the menu price of \emph{a} food item?} and \say{What is the menu price of \emph{my} food item?}, while the former is an \emph{I=Info} query to understand the definition of menu price, the latter is to know the existing menu price of user's food item which needs to fetch the details from a tool and hence should be classified as \emph{I=Action}.
    \item Multi-turn Conversation understanding: User can ask an action query and switch to informational query in the middle or vice-versa. The follow-up user messages can be partial and derive from the conversation context heavily (e.g., \say{What does \emph{this} mean?}). This requires IC to have nuanced conversation understanding to classify user message accurately.
    \item Handling domain specific acronyms: Conversation and tasks can refer to internal keywords and acronyms not present in common language usage. Knowledge of these are required to understand the context of conversation to act on it accurately.
    \item Context length: Conversations can be lengthy and run into several hundreds of tokens. Classifier needs to account for the complete context to make decisions.
\end{enumerate}
Table~\ref{tab:intent_classifier_prompting_techniques} provides a comprehensive comparative analysis of the effectiveness of each prompting technique.
Initially, we established a zero-shot prompt as our baseline, achieving an accuracy of 89.26\% with a latency of approximately 3 seconds using the \emph{claude-instant-v1} model. Subsequently, we investigated the efficacy of chain-of-thought prompting. This method involved presenting a sequence of yes/no questions within the same prompt to steer the Intent Classifier towards the accurate intent selection. (An illustrative example from the prompt is as follows: \say{\textit{Is the context directly related to digital restaurant platform or business? If Yes , Go to next step, If no Intent = Out of Context, Is the User asking the meaning or definition of retail terminologies?, If Yes, Intent = Information, If No, Go to next step}}). Despite its implementation, this prompting technique yielded a negligible uplift of less than 0.5\% in accuracy.
Another approach explored was one-vs-all prompting. Herein, we explicitly defined one intent (e.g. \emph{I=Info}) while categorizing the remainder as another intent. This technique proved efficient in mitigating ambiguity in the instructions, consequently yielding a 2\% improvement from the original baseline. Furthermore, by formulating a prompt with explicit instructions and examples for ambiguous scenarios (few shot prompting), we achieved the most significant enhancement thus far, with a 5\% uplift from the baseline performance.

In another experiment, we evaluated the performance of various instruct-tuned large language models (LLMs), the outcomes of which are delineated in Table~\ref{tab:intent_classifier_llm_compare}. The \emph{claude-3-sonnet} model emerged with the highest accuracy slightly exceeding 94\%, whereas the Mixtral model exhibited superior latency measures fine-tuning which will be a future work for improved accuracy. 

\subsection{LLM Agents Input Prompts \& Output Formatting }~\label{appendix:llm_input_output_formatting}
In this section we go deeper into the details of how we prompt our Task-Agents (LLM Agents in MARS) to get desired reasoning and output.

\textbf{LLM Input Prompt:} Below mention is a sample LLM Agent's prompt using which we intialise all our Task-Agents where details like \emph{agent\_name, agent\_purpose, agent\_task\_execution\_steps, sub\_task\_agents, tools, history, user\_message} are dynamic variables replaced with the actual values on the fly using Agent's internal state. We employ techniques like Chain-of-thought reasoning, guiding LLM to complete the prefix string (\emph{[Agent]<thinking>}) so that it steers in the required direction, output formatting instructions and XML tags to define segments in the prompts carefully.

\begin{lstlisting}
{{agent_name}}, {{agent_purpose}}
<TEP_STEPS>
{{agent_task_execution_steps}}
</TEP_STEPS>
Sub-Tasks:
<sub_tasks>
{{sub_task_agents}}
</sub_tasks>
Tools:
<tools>
{{agent_tools}}
</tools>
Place to Add important instructions:
<instructions>
{{instructions}}
</instructions>
Placeholder for chat history
<history> {{history}} </history>
\end{lstlisting}

\textbf{LLM Output:} We prompt the LLM to generate the following output format, which is then parsed to get relevant actions:
\begin{lstlisting}
<response>{
  "content": "The message to be conveyed back to the user.",
  "function_call": {
    "name": "function name",
    "arguments": "{\"Arg1\": \"Arg1_value\"}"
  }
}</response>
\end{lstlisting}

\subsection{LLM Evaluation Prompt}~\label{appendix:mars_llm_evaluation_prompt}
In this section we detail the LLM based semantic similarity matching LLM prompt for evaluating MARS Agents' responses. While verifying the generated function call and corresponding parameters is easy as they can be matched after parsing from the string with the ground truth deterministically, it can be challenging to match whether the LLM generated response back to the \emph{Actor/User} is same as the intended string in ground truth test set. Traditionally a manual audit is conducted to look at the generated string and ground truth string to identify if both have the same semantics or meaning. This can be a time taking and costly task depending on the size of your test dataset. We employ an LLM based task evaluation strategy where we prompt \emph{claude-instant-v1} to evaluate if two responses (sentence1 and sentence2) are semantically same or not. We conducted a manual audit as well and found a Cohen's Kappa score of 0.65 (96.66\% agreement) between auditors and LLM generated evaluations establishing the effectiveness of our approach. 




\subsection{Digital Restaurant Service Platform Conversation Dataset}~\label{appendix:retail_conv_dataset}
Each usecase has their own set of task execution procedure (TEP) steps in natural language, deterministic multi-step execution task and utility queries. 
Deterministic tasks (functions) are defined as JSONSchemas to the LLM prompt as input. A sample of TEP steps and a function JSONSchema is mentioned below:

\textbf{Sample Function JSONSchema for Restaurant Menu Update:}
\begin{lstlisting}
{
    "name": "menu_price_update_task",
    "description": "update the price for a menu item of a restaurant",
    "parameters": {
      "type": "object",
      "properties": {
        "merchant_id": {
          "type": "string",
          "description": "Unique identifier for a merchant"
        },
        "restaurant_name": {
          "type": "string",
          "description": "name of the restaurant"
        },
        "current_price": {
          "type": "string",
          "description": "current price of the menu item"
        },
        "new_price": {
          "type": "number",
          "description": "new price to be updated for the menu item"
        },
        "item_name": {
          "type": "string",
          "description": "name of the menu item for which the price needs to be updated"
        }
      },
      "required": [
        "merchant_id",
        "restaurant_name",
        "current_price",
        "new_price",
        "item_name"
      ]
    }
}
\end{lstlisting}

%% file: content/algorithms/guardrails_algorithm.tex


\newcommand\mycommfont[1]{\footnotesize\ttfamily\textcolor{blue}{#1}}
\SetCommentSty{mycommfont}

\SetKwInput{KwInput}{Input}                
\SetKwInput{KwOutput}{Output}              

\begin{algorithm}[!ht]
\SetNoFillComment
\DontPrintSemicolon
\caption{MARCO Reflection Guardrails}
\label{algo:marco_reflection_guardrails}
  
  \KwInput{
    $F_*^x = \{F_1^x, F_2^x, ..., F_n^x\}$, $P_*^x = \{P_{F_1^x}, P_{F_2^x}, ..., P_{F_n^x}\}$, \tcc{list of available tools, Sub-Agents \& respective parameters in \emph{Agent}$_x$} \\ 
    $R$ \tcc{LLM Agent generated response string}
  }
  \KwOutput{
    \emph{Agent}$_x$ updated context 
  }
  \If{invalid\_output\_format($R$)}{
        \emph{Agent}$_x$.add\_to\_context(\say{Output $R$ is not as per required formatting guidelines.})
        
  }
  \Else{
    $\hat{F}_i^x$, $\hat{P}_{F_i^x} \gets$ parse\_llm\_response($R$) \tcc{LLM generated Function \& corresponding parameters}
    \If{$\hat{F}_i^x \not\in F_*^x$}{
        \emph{Agent}$_x$.add\_to\_context(\say{Function $F_i^x$ not present in Agent tools and Sub-Agents.})
        
    }
    \For{$p \in \hat{P}_{F_i^x}$}{
        \If{$p \not\in P_{F_i^x}$}{
            $\hat{P}_{F_i^x} \gets \hat{P}_{F_i^x} \setminus p$ \tcc{remove $p$ from generated parameters set}
        }
        \ElseIf{$p$.value() $\not\in$ \emph{Agent}$_x$.user\_messages()}{
        \tcc{parameter value not present (grounded) in user messages}
            \emph{Agent}$_x$.add\_to\_context(\say{Value of $p$ not provided by the user.}) 
        }
        \Else{
            \emph{rules} $\gets$ get\_predefined\_rules\_errors($p$) \tcc{example \say{length($p$) should be $\leq$ 10}}
            \emph{Agent}$_x$.add\_to\_context(\say{Following rules not satisfied by $p$: \emph{rules}.})
        }
    }
  }
\end{algorithm}

%% file: content/tables/marco_reflection_ablation.tex
\begin{table*}[htbp]
\centering
\resizebox{\textwidth}{!}{%
\begin{tabular}{@{}lcccccccccccc@{}}
\toprule
\multicolumn{13}{c}{DRSP-Conv dataset}                                                                                                                                                                                                                                                                                                                                                                                                                                                                                                                                                                                                                                                                                                                                                                                      \\
\textbf{}             & \multicolumn{2}{c}{With All Reflections}                                                                                       & \multicolumn{2}{c}{\begin{tabular}[c]{@{}c@{}}Without Incorrect Formatting\\ Reflection\end{tabular}}                          & \multicolumn{2}{c}{\begin{tabular}[c]{@{}c@{}}Without Function Hallucination\\ Reflection\end{tabular}}                        & \multicolumn{2}{c}{\begin{tabular}[c]{@{}c@{}}Without Parameter Grounding\\ Reflection\end{tabular}}                           & \multicolumn{2}{c}{\begin{tabular}[c]{@{}c@{}}Without Parameter Static Rules \\ Reflection\end{tabular}}                       & \multicolumn{2}{c}{\begin{tabular}[c]{@{}c@{}}Without A Reflection \\ (retries = 0)\end{tabular}}                              \\ \cmidrule(lr){2-3} \cmidrule(lr){6-7} \cmidrule(lr){10-11}
Model Name            & \begin{tabular}[c]{@{}c@{}}Accuracy (\%) \\ ± Std dev\end{tabular} & \begin{tabular}[c]{@{}c@{}}Latency \\ (secs)\end{tabular} & \begin{tabular}[c]{@{}c@{}}Accuracy (\%) \\ ± Std dev\end{tabular} & \begin{tabular}[c]{@{}c@{}}Latency \\ (secs)\end{tabular} & \begin{tabular}[c]{@{}c@{}}Accuracy (\%) \\ ± Std dev\end{tabular} & \begin{tabular}[c]{@{}c@{}}Latency \\ (secs)\end{tabular} & \begin{tabular}[c]{@{}c@{}}Accuracy (\%) \\ ± Std dev\end{tabular} & \begin{tabular}[c]{@{}c@{}}Latency \\ (secs)\end{tabular} & \begin{tabular}[c]{@{}c@{}}Accuracy (\%) \\ ± Std dev\end{tabular} & \begin{tabular}[c]{@{}c@{}}Latency \\ (secs)\end{tabular} & \begin{tabular}[c]{@{}c@{}}Accuracy (\%) \\ ± Std dev\end{tabular} & \begin{tabular}[c]{@{}c@{}}Latency \\ (secs)\end{tabular} \\ \midrule
llama-3-8b-instruct   & 42.44 ± 2.01                                                       & 3.75                                                      & 16.29 ± 0.64                                                       & 4.38                                                      & 41.18 ± 0.96                                                       & 3.72                                                      & 41.09 ± 1.41                                                       & 3.7                                                       & 41.18 ± 1.2                                                        & 3.73                                                      & 15.93 ± 0.98                                                       & \textbf{1.9}                                              \\
mistral-7b-instruct   & 66.33 ± 1.04                                                       & 4.92                                                      & 64.52 ± 0.61                                                       & 5.17                                                      & 66.88 ± 1.3                                                        & 5.01                                                      & 64.34 ± 0.81                                                       & 5.02                                                      & 65.79 ± 0.52                                                       & 5.16                                                      & 59.28 ± 1.06                                                       & 2.9                                                       \\
mixtral-8x7b-instruct & 40.64 ± 1.51                                                       & 17.77                                                     & 39.46 ± 1.08                                                       & 20.42                                                     & 41.54 ± 0.98                                                       & 24.15                                                     & 40.82 ± 2.43                                                       & 23.61                                                     & 40.81 ± 1.26                                                       & 20.93                                                     & 32.67 ± 0.38                                                       & 15.55                                                     \\
claude-instant-v1     & 74.38 ± 1.4                                                        & 3.25                                                      & 72.5 ± 1.4                                                         & 3.37                                                      & 74.38 ± 1.4                                                        & 3.14                                                      & 74.38 ± 2.61                                                       & 2.85                                                      & 75.0 ± 0.0                                                         & 2.9                                                       & 53.12 ± 3.83                                                       & 2.53                                                      \\
claude-3-haiku        & 84.8 ± 0.88                                                        & \textbf{2.14}                                             & \textbf{84.43 ± 1.65}                                              & \textbf{2.13}                                             & 83.98 ± 1.3                                                        & \textbf{2.54}                                             & 78.73 ± 0.78                                                       & \textbf{2.37}                                             & 81.09 ± 1.08                                                       & \textbf{2.25}                                             & \textbf{75.2 ± 0.87}                                               & 2.24                                                      \\
claude-v2.1           & 88.51 ± 0.76                                                       & 8.44                                                      & 68.42 ± 1.34                                                       & 9.49                                                      & 88.42 ± 0.68                                                       & 8.22                                                      & 86.24 ± 1.09                                                       & 8.35                                                      & 86.15 ± 1.45                                                       & 8.19                                                      & 64.52 ± 1.04                                                       & 6.61                                                      \\
claude-3-sonnet       & \textbf{94.48 ± 0.59}                                              & 5.61                                                      & 73.39 ± 0.5                                                        & 6.23                                                      & \textbf{94.03 ± 0.59}                                              & 5.41                                                      & \textbf{91.04 ± 1.08}                                              & 5.5                                                       & \textbf{91.86 ± 0.46}                                              & 5.26                                                      & 66.34 ± 0.82                                                       & 4.07                                                      \\ \midrule
\multicolumn{13}{c}{Retail-Conv dataset}                                                                                                                                                                                                                                                                                                                                                                                                                                                                                                                                                                                                                                                                                                                                                                                    \\
\textbf{}             & \multicolumn{2}{c}{With All Reflections}                                                                                       & \multicolumn{2}{c}{\begin{tabular}[c]{@{}c@{}}Without Incorrect Formatting\\ Reflection\end{tabular}}                          & \multicolumn{2}{c}{\begin{tabular}[c]{@{}c@{}}Without Function Hallucination\\ Reflection\end{tabular}}                        & \multicolumn{2}{c}{\begin{tabular}[c]{@{}c@{}}Without Parameter Grounding\\ Reflection\end{tabular}}                           & \multicolumn{2}{c}{\begin{tabular}[c]{@{}c@{}}Without Parameter Static Rules \\ Reflection\end{tabular}}                       & \multicolumn{2}{c}{\begin{tabular}[c]{@{}c@{}}Without A Reflection \\ (retries = 0)\end{tabular}}                              \\ \cmidrule(lr){2-3} \cmidrule(lr){6-7} \cmidrule(lr){10-11}
Model Name            & \begin{tabular}[c]{@{}c@{}}Accuracy (\%) \\ ± Std dev\end{tabular} & \begin{tabular}[c]{@{}c@{}}Latency \\ (secs)\end{tabular} & \begin{tabular}[c]{@{}c@{}}Accuracy (\%) \\ ± Std dev\end{tabular} & \begin{tabular}[c]{@{}c@{}}Latency \\ (secs)\end{tabular} & \begin{tabular}[c]{@{}c@{}}Accuracy (\%) \\ ± Std dev\end{tabular} & \begin{tabular}[c]{@{}c@{}}Latency \\ (secs)\end{tabular} & \begin{tabular}[c]{@{}c@{}}Accuracy (\%) \\ ± Std dev\end{tabular} & \begin{tabular}[c]{@{}c@{}}Latency \\ (secs)\end{tabular} & \begin{tabular}[c]{@{}c@{}}Accuracy (\%) \\ ± Std dev\end{tabular} & \begin{tabular}[c]{@{}c@{}}Latency \\ (secs)\end{tabular} & \begin{tabular}[c]{@{}c@{}}Accuracy (\%) \\ ± Std dev\end{tabular} & \begin{tabular}[c]{@{}c@{}}Latency \\ (secs)\end{tabular} \\ \midrule
llama-3-8b-instruct   & 49.68 ± 1.55                                                       & 3.44                                                      & 20.32 ± 0.46                                                       & 4.99                                                      & 48.47 ± 0.53                                                       & 3.45                                                      & 46.77 ± 1.76                                                       & 3.41                                                      & 47.58 ± 1.18                                                       & 3.36                                                      & 17.82 ± 1.12                                                       & \textbf{1.64}                                             \\
mistral-7b-instruct   & 55.32 ± 0.77                                                       & 4.89                                                      & 54.68 ± 1.29                                                       & 4.96                                                      & 54.03 ± 0.57                                                       & 4.55                                                      & 55.16 ± 1.04                                                       & 4.74                                                      & 55.24 ± 1.56                                                       & 4.82                                                      & 50.72 ± 0.66                                                       & 3.06                                                      \\
mixtral-8x7b-instruct & 48.31 ± 0.60                                                       & 12.94                                                     & 47.34 ± 1.5                                                        & 11.82                                                     & 48.87 ± 2.18                                                       & 10.35                                                     & 50.32 ± 1.75                                                       & 11.24                                                     & 48.87 ± 2.82                                                       & 10.22                                                     & 40.49 ± 0.93                                                       & 5.96                                                      \\
claude-instant-v1     & 76.61 ± 0.81                                                       & 4.14                                                      & 68.95 ± 0.57                                                       & 4.32                                                      & 75.56 ± 0.61                                                       & 4.23                                                      & 60.56 ± 0.34                                                       & \textbf{2.94}                                             & 74.03 ± 1.05                                                       & 4.28                                                      & 60.56 ± 0.24                                                       & 2.94                                                      \\
claude-3-haiku        & 87.82 ± 0.44                                                       & \textbf{2.45}                                             & 87.58 ± 0.78                                                       & \textbf{2.29}                                             & 86.21 ± 1.47                                                       & \textbf{2.28}                                             & 82.74 ± 0.33                                                       & 3.1                                                       & 85.08 ± 0.57                                                       & \textbf{3}                                                & 77.66 ± 1.01                                                       & 2.43                                                      \\
claude-v2.1           & 92.34 ± 0.49                                                       & 8.2                                                       & \textbf{88.31 ± 0.57}                                              & 8.6                                                       & 90.32 ± 1.14                                                       & 6.22                                                      & \textbf{87.98 ± 0.44}                                              & 8.63                                                      & 89.68 ± 0.67                                                       & 8.48                                                      & \textbf{78.87 ± 0.61}                                              & 6.95                                                      \\
claude-3-sonnet       & \textbf{92.74 ± 0.49}                                              & 5.85                                                      & 66.53 ± 0.81                                                       & 7.93                                                      & \textbf{91.53 ± 0.64}                                              & 8.55                                                      & 83.39 ± 2.91                                                       & 6.61                                                      & \textbf{90.89 ± 0.54}                                              & 6.16                                                      & 60.89 ± 0.81                                                       & 4.61                                                      \\ \bottomrule
\end{tabular}%
}
\caption{LLMs performance comparison for MARCO by removing different type of reflection guardrails on DRSP-Conv and Retail-Conv datasets averaged across 5 runs.}
\label{tab:marco_reflection_ablation}
\end{table*}

%% file: content/figures/mars_temperature_effect.tex
\begin{figure}[htbp]
  \centering
  \includegraphics[width=\columnwidth]{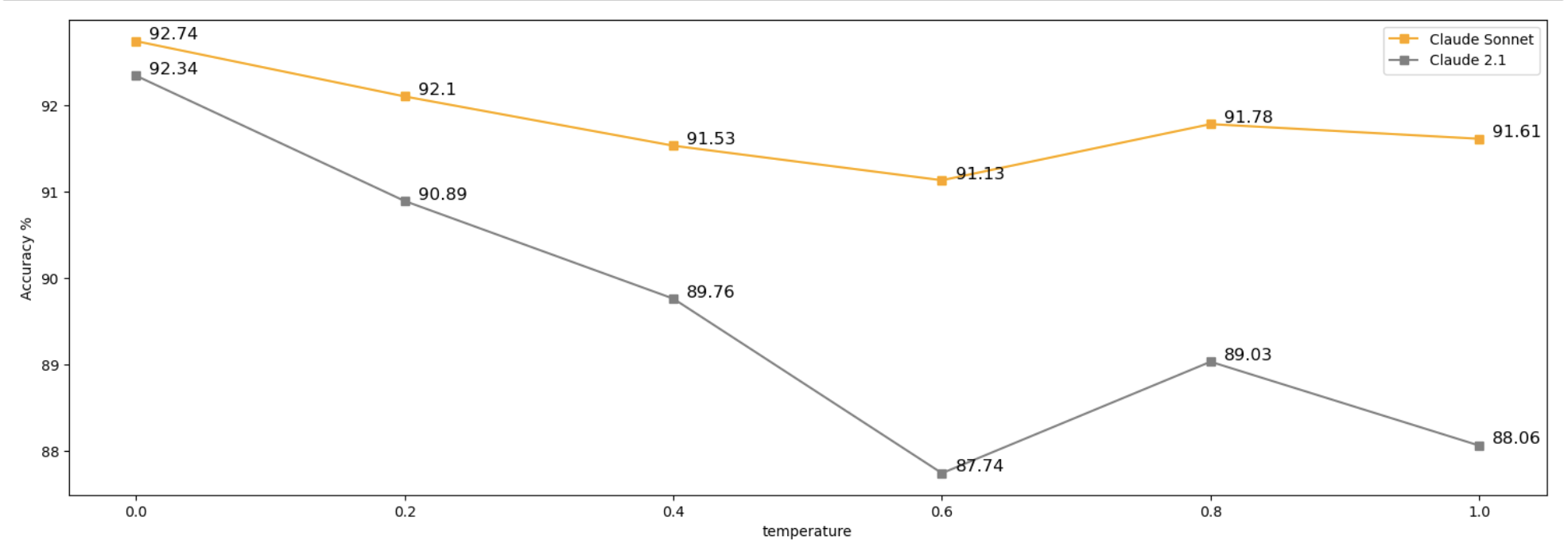}
  \caption{Effect of temperature hyper-parameter on MARS performance.}
  \label{fig:mars_temperature_effect}
\end{figure}

%% file: content/figures/mars_input_output_token_latency_relation.tex
\begin{figure}[htbp]
  \centering
  \includegraphics[width=\columnwidth]{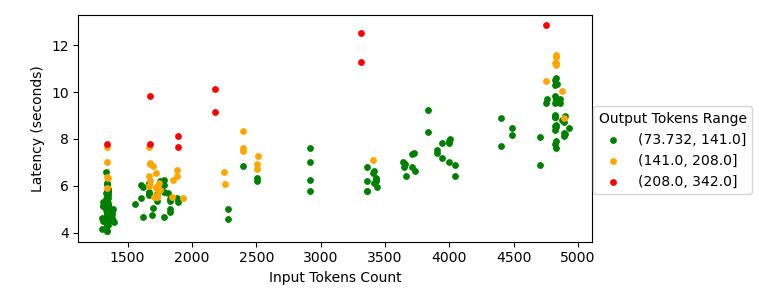}
  \caption{Correlation between number of input and output tokens in LLM prompt and response latency for MARS using \emph{claude-3-sonnet}.}
  \label{fig:mars_input_output_token_latency_relation}
\end{figure}

%% file: content/tables/intent_classifier_prompting_techniques.tex
\begin{table}[htbp]
\centering
\resizebox{0.85\columnwidth}{!}{%
\begin{tabular}{@{}lcc@{}}
\toprule
Prompting Technique & Average Accuracy (\%) $\pm$ Std dev & Average Latency \\ \midrule
Zero Shot           & 89.26\% $\pm$ 0.47                  & 2.99            \\
Chain of Thought    & 89.68\% $\pm$ 1.56                  & \textbf{1.98}   \\
One Vs. All         & 91.37\% $\pm$ 0.47                  & \textbf{1.98}   \\
Few Shot            & \textbf{94.32\% $\pm$ 0.94}         & 2.43            \\ \bottomrule
\end{tabular}%
}
\caption{Intent Classifier performance comparison based on varying prompting techniques.}
\label{tab:intent_classifier_prompting_techniques}
\end{table}

%% file: content/tables/intent_classifier_llm_compare.tex
\begin{table}[htbp]
\centering
\resizebox{0.9\columnwidth}{!}{%
\begin{tabular}{@{}lcc@{}}
\toprule
Model Name            & Average Accuracy (\%) $\pm$ Std dev & Average Latency \\ \midrule
mixtral-8x7b-instruct & 65.47\% $\pm$ 0.008                 & \textbf{1.62}   \\
mistral-7b-instruct   & 75.58\% $\pm$ 0.004                 & 1.96            \\
claude-3-haiku        & 90.32\% $\pm$ 0.88                  & 1.98            \\
claude-v2.1           & 92.42\% $\pm$ 0.47                  & 5.02            \\
claude-instant-v1     & 94.32\% $\pm$ 0.94                  & 2.43            \\
claude-3-sonnet       & \textbf{94.53\% $\pm$ 0.88}         & 1.98            \\ \bottomrule
\end{tabular}%
}
\caption{Comparing Intent Classifier performance and latency using various LLMs.}
\label{tab:intent_classifier_llm_compare}
\end{table}